# Extracting Semantics from Maintenance Records

**Sharad Dixit, Varish Mulwad, Abhinav Saxena**
GE Research
{sharad.dixit, varish.mulwad, asaxena}@ge.com

## Abstract

Rapid progress in natural language processing has led to its utilization in a variety of industrial and enterprise settings, including in its use for information extraction, specifically named entity recognition and relation extraction, from documents such as engineering manuals and field maintenance reports. While named entity recognition is a well-studied problem, existing state-of-the-art approaches require large labelled datasets which are hard to acquire for sensitive data such as maintenance records. Further, industrial domain experts tend to distrust results from black box machine learning models, especially when the extracted information is used in downstream predictive maintenance analytics. We overcome these challenges by developing three approaches built on the foundation of domain expert knowledge captured in dictionaries and ontologies. We develop a syntactic and semantic rules-based approach and an approach leveraging a pre-trained language model, fine-tuned for a question-answering task on top of our base dictionary lookup to extract entities of interest from maintenance records. We also develop a preliminary ontology to represent and capture the semantics of maintenance records. Our evaluations on a real-world aviation maintenance records dataset show promising results and help identify challenges specific to named entity recognition in the context of noisy industrial data.

## 1 Introduction

Within industrial companies, large unlabeled text datasets often come in the form of engineering manuals, design documents, and field maintenance reports. For instance, in the aviation domain, relevant data are captured in maintenance logs, work orders, post flight reports, delays/cancellation data, etc. Maintenance (MX) reports and work orders can be particularly valuable because they describe issues that a field engineer or a technician observes with an asset and informal instructions on how the engineer was (or was not) able to solve that issue. Information extraction, specifically entity and relation extraction, from tens to hundreds of thousands of unstructured textual maintenance records across a fleet of a large industrial asset (such as a specific gas turbine, aircraft engine, airframe or a nuclear power plant) can derive tremendous value, including information on what are frequent and rare failure modes, root cause for each failure, and which resolutions effectively address each type of failure. Entities and relations of interest for such use-cases include the affected components/parts (e.g., *gasket*) referenced in the text, location (e.g. *intake*), along with observations (e.g., *leaking*), actions performed (e.g., *replaced*) and the relation between them (e.g. *<intake gasket> <hasObservation> <leaking>*). Semantic Web technologies and Knowledge Graphs are a natural choice to capture this information in the form of structured semantics.

MX records, even though mostly unstructured, are timestamped and offer a valuable proposition of attaching contextual history to incredible amounts of time series data that are collected from these assets. There have been significant advances in modeling time series for anomaly detection and diagnosis of faults, however these data are hard to label with specific actions or events that the asset may have experienced. Semantic extraction of root cause, failure, part replacement, repair or anything related to asset's history offers a means to automatically label time series data for modeling (supervised) and validation (both supervised and unsupervised) of machine learning/deep learning (MLDL) models. This can save valuable time of domain subject matter experts (SMEs) from excruciating tasks of manually generating the context looking through the logs and providing labels to model developers. Furthermore, automating this task also removes subjective variability from person to person and allows for more consistent data processing and performance benchmarking. MX logs embed a wealth of knowledge that is crucial to explain observed effects in time series data, and hence if appropriately combined through semantics add explainability to MLDL outputs, which is a key metric for deploying such models for critical industrial assets such as jet engines and power gas turbines. Traditionally time series analysis focuses on detection and identification of failure mode, however semantic extraction allows us to label the time series not just with a failure condition but also contextually relevant part action (replacement/repair). This allows for developing prescriptive models (data to action prediction), outputs of which could be valuable in logistics and inventory planning for optimizing maintenance schedules and minimizing asset downtimes.

Yet another application of semantic extraction is in analyzing the effectiveness of maintenance actions. In heavily regulated industries like aviation and nuclear, preventive maintenance or the concept of life limited parts (LLPs) is mandated, where preventive maintenance must be performed on a predetermined schedule. By analyzing logs or workorder to determine which preventive maintenance actions resulted in effective repair versus those resulted in subsequent corrective maintenance activity we can potentially improve maintenance processes. This also allows for establishing better ground truth in evaluating whether time series data-driven monitoring could have pointed to correct diagnosis to suggest correct repair actions in the first place. Furthermore, in domains like nuclear plants, many systems are not monitored through sensors in the first place and the only source of monitoring information is though manual inspection reports and operator observations. Semantic extraction of these data offers means to developing diagnostics and prognostics models in such scenarios.

State-of-the-art (SOTA) entity and relation extraction approaches use supervised deep neural network architectures requiring large labelled datasets. The acquisition of such large and diverse labelled datasets remains a key challenge in an industrial setting. Data confidentiality and lack of domain expertise (outside of organizations) makes it infeasible to use systems such as Amazon Mechanical Turk to obtain inexpensive annotations. Further, industrial domain experts tend to distrust results from black box machine learning models, especially when such extracted semantics are used in downstream predictive maintenance analytics. We address these challenges by developing approaches for extracting and capturing semantics from MX records comprising of entity and relation extraction techniques that combine ontology-driven extraction with pre-trained language models. In our experience, experts tend to put more faith in domain knowledge guided ontology-driven extraction processes. Dictionary approaches tend to face the challenge of scale, but some of those may not impact much when processing MX records. Entities belonging to classes such as **Observation** and **Action** are generally limited in number to be captured in an ontology. **Components/Parts**, while larger in number, are still finite and can also be captured as domain knowledge in an ontology. A challenge with components is that technicians tend to use many variations while referring to them and/or include additional contextual information (e.g., *left engine #4 cylinder baffle cracked*). Source of these variations ranges from geographic locations, cultures, languages, as well as processes across different operator organizations (e.g. different airlines or utility companies). We handle such issues and the extraction of additional context along with the component entities by exploring both the usage of domain rules combining syntactic and semantic information with the use additional dictionaries and pre-trained language models.

The key contributions of this paper include: i) an preliminary ontology for representing the semantics extracted from MX records, ii) an enhanced rule-based approach which leverages domain ontologies along with syntactic and semantic knowledge for entity extraction, iii) an Question Answering (QA) approach for entity and relation extraction, and iv) evaluation on a real world aviation MX records dataset comparing the base dictionary, the enhanced rule-based, and the QA model approaches. While the rest of paper references running examples for aviation MX records, we posit that these approaches can generalize to MX records for assets from other industries such as Gas and Nuclear Power.

## 2 Approach

We begin by describing the structure of a MX record and a preliminary ontology developed to represent the extracted semantics. We then describe three different approaches for entity (and relation) extraction from MX records.

### 2.1 Anatomy of a MX Record

A textual MX record in the aviation domain provides key information about affected parts (*wire*), their associated components (*landing gear actuator*), along with observations (*leaking*), actions performed (*replaced*), and component location (*left outboard*) especially when multiple similar systems exist (e.g. four jet engines on an aircraft). Additionally, it references specific sections from a maintenance manual that a technician may have referred to resolve an observed issue. It also references any tests that they have performed to verify the fix and report the results of the test (e.g., *performed ground test, check good.*). A typical MX record comprises of a paragraph of several brief sentences, with each sentence conveying different technical information. For instance, the first sentence may describe the affected component and the observed problem. The next one may talk about the action(s) performed on the affected component, so on and so forth. For the purposes of this paper, we use an off-the-shelf model to divide paragraph into sentences and extract semantics at the sentence level. For rest of the paper, when we reference "MX record", it refers to a single sentence from a MX record paragraph.

Entities of interest that can help capture the semantics of a MX record include **Part**, **Component**, **Observation**, and **Action**. Relations of interest to capture the semantics include "**has associated observation**" between Part/Component and Observation and "**has associated action**" between Part/Component" and Action. Components represent high level systems/sub-systems in an aircraft, whereas Part represents typical parts present in a system/sub-system. Parts and Components can be associated with additional contextual information such as its **Ordinal** (*#1 intake gasket*) and **Location** (*top left baffle*). We represent the semantics of MX records (as described below) by extracting entities for all the referenced types. Our current work doesn't distinguish between a Part and Component; rather it extracts as one large phrase i.e. instead of tagging *motor* as [Part] and *brake* as [Component], we tag *motor brake* as [Component].

### 2.2 Representing the Semantics of MX Records

```
uri "http://mxrecords/".

MaintenanceRecord is a class
    described by recordId with a single value of type string,
    described by assetId with a single value of type string,
    described by dateActivityPerformed with a single value of type dateTime,
    described by maintenanceActivity with values of type MaintenanceActivity.

MaintenanceActivity is a class,
    described by hasAssociatedComponentOrPart with a single value of type ComponentOrPart.

ComponentOrPart is a class,
    described by hasName with a single value of type string,
    described by hasAssociatedOrdinal with a single value of type int,
    described by hasAssociatedLocation with a single value of type string,
    described by hasAssociatedObservation with values of type string,
    described by hasAssociatedAction with values of type string.
```

*Figure 1 A snapshot our Maintenance Records OWL ontology serialized using the Semantic Application Design Language (Crapo and Moitra, 2013)*

We develop an OWL (OWL Working Group, 2009) ontology to represent the semantics extracted from MX records. Maintenance records are represented by the **MaintenanceRecord** class in our ontology. Properties such as **recordId**, **assetId** (an identifier for the asset, e.g. an airframe or engine, with which the record is associated), **dateActivityPerformed** capture key metadata information associated with the maintenance records. Many such attributes can be associated, but we highlight a few for the purposes of brevity. Every maintenance record can be associated with multiple maintenance activities, each activity extracted from an individual sentence in the record. The activities described in the sentence are captured via the **MaintenanceActivity** class. A sentence typically describes the affected part, component, along with the associated observation and actions performed on them. The part/component may also have additional information such as its ordinal (e.g. 1$^{st}$ v/s. 2$^{nd}$) and location (e.g. left, right, etc.). **MaintenanceActivity** is linked to the affected component via the **hasAssociatedComponentOrPart** property with the range **ComponentOrPart**. The **ComponentOrPart** class captures extracted part/component name (**hasName)**, observation (**hasAssociatedObservation**), action (**hasAssociatedAction**), and the context associated with the component such as its ordinal (**hasAssociatedOrdinal**) and location (**hasAssociatedLocation**). A snapshot of our ontology can be seen in Figure 1.

### 2.3 Ontology-driven Base Approach

Our base approach leverages a dictionary driven UIMA ConceptMapper (Tanenblatt, Coden, and Sominsky, 2010) to extract entities of interests from text. Over the years, we and our colleagues with aviation asset maintenance expertise have developed internal domain dictionaries (both for the engine and airframe) capturing lists of Components, Positions, Observations, and Actions. Every term in the domain dictionary is associated with its synonyms and a few spelling variations. These domain dictionaries have been developed in the form of OWL ontologies by mapping every unique term as a class and its synonyms and spelling variations as the RDFS[1] labels of the class. We automatically translate the domain term ontologies into UIMA ConceptMapper dictionaries in XML format. Each class is translated into its canonical form and the variants for the dictionary format. The canonical form is represented by a uniform resource identifier of the class, while its variants include the preferred name, synonyms, and spelling variations. Given a sentence, our base approach returns a list of extracted Components, Positions, Observations, and Actions. Each extracted entity includes the text string, text span (start and end positions) and its semantic type (e.g., Action).

### 2.4 Syntactic and Semantic Rule-based Approach

Some of the challenges with a ontology-driven approach, especially for component and location extraction, include several variations used by technicians in text, entity names broken into smaller sub entities in the ontology (e.g., *flap actuator* represented as separate entities *flap* and *actuator*) and overlapping entities for representing both coarse-grained and fine-grained instances in the ontology (e.g. *seal* and *baffle seal*). Additionally, while component names are captured in the ontology/domain dictionaries, typical parts (e.g., *nuts*, *valves*, *wires*, *bearing*, *bush*, *motor*, etc.) are not. Often the observation or the action referenced in text is associated with a part within a component as opposed to the entire component itself. This information is crucial for downstream systems consuming the extracted semantics.

Our rule-based approach overcomes these challenges and extracts additional contextual information such as typical parts by leveraging both syntactic and semantic clues associated with the entities extracted by the base approach. The base approach returns both the coarse and fine-grained entities. The **first rule** in this approach extracts fine-grained and discards coarser entities. It does so by comparing the span of the entities that have the same start position and retaining the ones with the longer span. Our observation and analysis of the terms in the domain ontologies show that fine-grained entities were longer in span as opposed to their coarser counterparts.

A **second generic rule** then loops over the reduced set of entities to identify if sub entities can be joined and extract additional contextual information. If there are two neighboring consecutive entities of the same semantic type with no gap between them (e.g., *flap* [Component] *actuator* [Component]), the rule joins them as a single entity (e.g. *flap actuator* [Component]).The rule also checks for entity pairs separated by a distance of less than or equal to *k* in the text. Useful contextual information, mainly typical parts, is present between such entity pairs. Consider a sub sequence of a tagged sentence: *removed* [Action] *motor brake* [Component], where action and component are extracted by the base approach, but typical part *motor* isn't. Since the span between action and component is small, motor is extracted and attached with the component name i.e. *brake* [Component] becomes *(motor)*

---
[1] https://www.w3.org/TR/rdf-schema/

| | Precision | | | Recall | | | F1 | | | | Precision | | | Recall | | | F1 | | |
|---|---|---|---|---|---|---|---|---|---|---|---|---|---|---|---|---|---|---|---|
| Entity Types | Base | Rule | QA | Base | Rule | QA | Base | Rule | QA | Entity Types | Base | Rule | QA | Base | Rule | QA | Base | Rule | QA |
| Action | 0.64 | **0.82** | 0.64 | **0.88** | 0.65 | **0.88** | **0.74** | 0.73 | **0.74** | Action | 0.67 | **0.86** | 0.67 | **0.92** | 0.68 | **0.92** | **0.77** | 0.76 | **0.77** |
| Component | 0.45 | **0.46** | 0.17 | **0.62** | 0.37 | 0.19 | **0.52** | 0.41 | 0.18 | Component | 0.68 | **0.71** | 0.41 | **0.94** | 0.57 | 0.45 | **0.79** | 0.63 | 0.43 |
| Observation | **0.24** | 0.12 | **0.24** | **0.30** | 0.05 | **0.30** | **0.27** | 0.07 | **0.27** | Observation | **0.25** | 0.24 | **0.25** | **0.31** | 0.10 | **0.31** | **0.28** | 0.14 | **0.28** |
| Location | 0.49 | **0.86** | 0.68 | **0.55** | 0.25 | 0.27 | **0.52** | 0.39 | 0.39 | Location | 0.77 | **0.93** | 0.81 | **0.85** | 0.27 | 0.32 | **0.81** | 0.42 | 0.46 |
| Ordinal | 0.00 | **1.00** | 0.96 | 0.00 | 0.22 | 0.32 | 0.00 | 0.36 | **0.47** | Ordinal | 0.00 | **1.00** | 0.96 | 0.00 | 0.22 | 0.32 | 0.00 | 0.36 | **0.47** |
| All | 0.50 | **0.62** | 0.41 | **0.62** | 0.41 | 0.43 | **0.56** | 0.49 | 0.42 | All | 0.65 | **0.76** | 0.53 | **0.80** | 0.50 | 0.55 | **0.72** | 0.61 | 0.54 |

(a)      (b)

*Figure 2 Evaluation metrics for (a) strict matching and (b) fuzzy matching with similarity score of 0.5*

*brake* [Component]. A parenthesis is added around typical part to indicate its nature as contextual information.

When looping over entities, the **second rule** is first applied to entities of type location. Once the location sub entities are joined as a single entity (e.g., joining independent locations *left* and *inboard* as *left inboard*), they are removed from the entity list and the rule is applied to the remaining set of entities (components, actions, and observations). The entities names are cleaned up by removing stop words.

While the key focus of this approach is entity extraction, we add a **third heuristic rule** to extract entity relations. The third rule checks for entity pairs that are adjacent or separated by less than or equal to *k* with semantic types Component-Action/Action-Component and Component-Observation/Observation-Component. This heuristic is thus able to extract the "hasAssociatedObservation" and "hasAssociatedAction" relationship between a component and its observation and action values for entity pairs that match this rule.

### 2.5 Question Answering (QA) Approach for Entity and Relation Extraction

We leverage part of our base approach and combine it with a pre-trained language model fine-tuned on a QA task to extract component/part names from a MX record (sentence). We frame the component/part entity extraction problem as answer span detection for a given question and context text. A MX record forms the context text, and a question is automatically generated based on the actions and observations extracted by the base approach. The generated question and the context text are provided as input to a pre-trained language model fine-tuned for a QA task. An overview of our approach is shown in Figure 3. Once an observation/action is extracted, a question of the form "What was [ACTION | OBSERVATION]?" is generated (e.g., "What was replaced?"). The response from the QA model is tagged as component/part. This not only extract components but also entity relations between component/part and observation/action. The extracted component/part name is further processed by the rule-based approach to identify references to ordinals and locations.

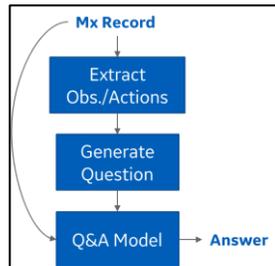

*Figure 3 Overview of the QA approach for entity and relation extraction*

## 3 Evaluation

We evaluate our approaches on a real-world English language aviation MX records dataset comprising over 110,000 unlabeled sentences. Two authors of this paper, after consultation with domain experts, annotated 578 randomly chosen sentences with Component, Action, Observation, Location, and Ordinal entity types. Part mentions were annotated as Components for these experiments. The annotations resulted in a total of **1065 entities**, with 434 Components, 332 Actions, 84 Observations, 142 Locations, and 73 Ordinals.

We compare the extracted entities with ground truth (gt) entities to compute precision, recall, and f-score (F1) for each of the approaches. Precision is computed as the number of correctly extracted entities divided by the total number of extracted entities. Recall is computed as the number of correctly extracted entities divided by the total number of expected entities. F-score is the harmonic mean of precision and recall. To determine the correctness of the extracted entity, its comparison with the gt entity was done in two ways (as is the case in the evaluation of named entity recognition systems; see section 5 in (Yadav and Bethard, 2018)): a) Strict match, ensuring both the both the entity boundary and the type match (i.e. extracted entity *landing gear actuator* with type Component is considered a correct extraction when gt entity is also *landing gear actuator* with type Component) and b) Fuzzy match, relaxing the constraint for entity boundary match, but still ensuring the entity type matches (i.e. extracted entity *gear actuator* with type Component could be considered a correct extraction when gt entity is *landing gear actuator* with type Component). We compute Sørensen–Dice similarity coefficient to compare gt and extracted entity strings. We compute metrics for similarity scores ranging from 0.5 to 1.0 (strict match) to study the effect of fuzzy entity boundary matching. DistilBERT (Sanh et al., 2019) fine-tuned on the Squad dataset (Rajpurkar et al. 2016) for used in the QA approach. The value of *k* was set to 10 in the Rule-based approach.

Figure 2 (a) shows the **strict entity boundary matching results** for the Base (Base), Rule-based (Rule), and the QA (QA) approaches. All the three approaches have reasonable success in extracting the Action entities. While the F1 score for all three approaches are similar, the base and QA approaches are better at extracting most of the expected actions (higher recalls), while the rule-based approach is more accurate at extracting actions (better precision). The reason for lower precision for Base and QA is due to the extraction of generic valid actions (e.g. *check*, *service*, *restored*) but not appearing as actions in the context of the MX records in our dataset (e.g. *aircraft ready to resume service*; *operations check good*). The rule-based approach eliminates such actions since its extracts them only in the context or neighborhood of extracted components. While this leads to a higher precision, it also results in a lower recall due to the presence of MX records only referencing actions without components or when component extraction fails.

The overall lower scores for extracting component entities can be, in part, attributed to the presence of ambiguous entities. Our domain dictionaries consist of generic entities such as *card (e.g., network card)* and *steps*, while the frequent context in which these terms appear in our dataset is the form of "*followed steps in reference card 123*." This impacted precision for the base and rule-based approaches. Base's better precision and recall scores can be attributed to the use of clean dictionary terms for a simple lookup. However, it does fail to extract typical parts associated with the components absent in the dictionaries. As described earlier, the rule-based approach relies on a distance $k$ ($k = 10$ in our experiments) between the entities extracted by the base approach, mainly to extract contextual information such as typical parts. Sources of error in this approach include extraction of incorrect context between entities (not referencing parts), unable to decompose components and locations accurately, and removal of stop words leading to entities at longer distance concatenated as a single component (which otherwise shouldn't be and are annotated as separate entities in gt). The QA approach relies on leveraging the inherent linguistic structures learnt by pre-trained language models (Manning et al., 2020) for the successful extraction of component names. However, poor linguistic structure in MX records led to the identification of incorrect text spans used for component extraction. Extracted text spans generally include way more additional contextual information along with the overall component name. While this work well in a standard QA setting, it hurts when used in the context of entity extraction. Incorrectly extracted action names also led to downstream subsequent incorrect component extraction since the wrong questions were posed to the QA model.

Lower scores for the Observation entity types can also be attributed to presence of ambiguous entities such as *stall*, *switch*, *ground* which can reference either an observation or a component/location. Several valid observation dictionary terms were also extracted by our approaches but were not deemed as observation entities in the gt by our annotators. Issues with the base approach for location entity extraction are similar to the ones with component extraction: incomplete entity extraction (e.g. extracting *r/h* instead of *r/h otbd*) and extracting dictionary terms that do not necessarily reference locations in the given context. Both rule-based and QA approach use a similar approach for location extraction and suffer from the same issues of inability to decompose location and component into appropriate entities from a larger contextual string and extraction/concatenation of much larger reference text. Challenges with ordinal extraction can be attributed to the inability of the regular expressions to identify and extract ordinal strings. Ordinal extraction operates on the component name string extracted in both rule-based and QA approaches. The longer text spans extracted by the QA approach helps capture the ordinal references; however, the component name strings extracted by the rule-based approach failed to capture ordinals as part of its contextual information.

Figure 2 (b) shows the results when the strict entity boundary matching is relaxed. Extracted entities are considered as correct whenever the Sørensen–Dice similarity coefficient between the extracted and gt entity is $>= 0.5$. The overall precision, recall, and F1 scores for the three approaches show improvements indicating the prediction of entity boundaries as a major challenge with all our current approaches. Significant improvements can be seen for Component and Location extraction. As described previously, the extraction for these entity types were most impacted by incorrect entity boundary prediction (e.g., incomplete entity extraction, additional context extraction, inability to accurately decompose components and locations etc.). Significant improvements can be seen for component extraction both for the base and rule-based approaches. The base approach though is missing crucial contextual information such as typical parts, which the rule-based and QA approaches extract, albeit, currently in a noisy fashion.

## 4 Related Work

Named entity recognition is a well-researched area. State-of-the-art systems are evaluated on typical entity types such as person, place, and organization (Yadav and Bethard 2018). Named Entity Recognition systems have also been developed for specific scientific domains such as medical (Xu et al., 2017), biomedical (Lee et al. 2019) and computer science (Beltagy, Lo, and Cohan, 2019) to name a few.

Extracting domain specific entities from MX records has received limited attention. (Chandramouli, Subramanian, and Bal, 2013) present an unsupervised method for extracting part names. Their method requires a set of input seed part names from which it generates candidate unigrams and bigrams to capture what we refer to as additional contextual information. Mutual information metric and part of speech tags are used to rank and filter the candidate part names across the corpus. While this helps discover all parts within a corpus, it doesn't help to extract and link them to individual MX records. (Niraula, Whyatt, and Kao, 2018) preset a semi-supervised iterative method, also, for extracting part names from MX records. Their method too requires a limited set of seed part names (they refer them as head nouns, e.g. *valve*, *gear*, etc.) which are used to generate variable length larger names with contextual information. These are further used as labels to train a conditional random field (CRF) model. They evaluated the quality of the extracted "head nouns" with subject matter experts rating each as "Good", "Borderline", or "Bad". 48% of the extracted head nouns were considered good or borderline demonstrating the challenges associated with extracting domain specific entities from MX records. (Niraula,

Whyatt, and Kao, 2020) further extend their previous work to extract both parts and conditions (i.e. what we refer to as observations). They manually annotated a dataset with parts and conditions and trained a bidirectional Long Short Term Memory (LSTM) model with a CRF at the top (architecture similar to the in Lample et al. 2016). Contrasting with existing works, our paper focuses on extracting comprehensive semantics from MX records with fine-grained entity types such as Parts, Components, Actions, Observations, Ordinals, and Locations with comparable results to current methods. We also include a preliminary ontology to represent the extracted semantics, thereby making it useful to be leveraged in downstream predictive MX analytics. Due to the sensitive and proprietary nature of such datasets, a direct comparison cannot be performed.

## 5 Lessons Learnt and Future Directions

We explored three approaches for extracting the semantics from MX records. The base ontology/dictionary driven approach shows promising results in cases when a comprehensive list of entities can be captured by domain experts in an unambiguous fashion. Such approaches work well, up to certain extent, for entity types such as Actions and Components. They however fail to capture useful and necessary contextual information such as typical parts. Our extended approaches developed on top of the base dictionary approaches are able to extract such contextual information, albeit in a rather noisy fashion, leading to over estimation of the entity boundaries. Accurate entity boundary prediction and cleaner separation or decomposition between parts, components, ordinals, and locations emerged as key problems to address to improve the overall quality of the extracted semantics. We were also able to demonstrate that off-the-shelf, pre-trained language models can be used (with limited success) on completely novel domains such as aviation MX records. With the challenge of limited labelled data, we will explore how our current approaches and the lessons learnt could be used to automatically label data akin to distant supervision (Mintz et al., 2009) and weak supervision (Ratner at al., 2017). Our base approach, with certain improvements, can be used to label MX records with entities such as Actions, Components, and Locations. We will investigate how the base approach, could be combined either with the rule-based or QA approach to decompose contextual component strings in Parts and Locations, such that the MX records can be annotated with those entity types. A weakly-labelled dataset then could be used to train a supervised model with SOTA neural network architectures for named entity recognition. Techniques such as (Virani, Iyer, and Yang, 2020) can be used to characterize the model's confidence in its prediction to gain trust of industrial domain experts.